\documentclass[letterpaper]{article} 
\usepackage{aaai23}  
\usepackage{times}  
\usepackage{helvet}  
\usepackage{courier}  
\usepackage[hyphens]{url}  
\usepackage{graphicx} 
\usepackage{natbib}  
\usepackage{caption} 
\usepackage[inline]{enumitem}
\usepackage[dvipsnames]{xcolor,colortbl}
\usepackage{makecell}
\usepackage{amsfonts}
\usepackage{amssymb}
\usepackage{float}

\usepackage{subcaption}

\usepackage{newfloat}
\usepackage{listings}
\usepackage{multirow}
\usepackage{booktabs}
\usepackage{times}
\usepackage[inline]{enumitem}
\usepackage{latexsym}
\usepackage[T1]{fontenc}    
\usepackage[utf8]{inputenc}

\usepackage[inline]{enumitem}
\usepackage[dvipsnames]{xcolor,colortbl}
\usepackage{amsmath,amssymb,amsfonts}
\usepackage[utf8]{inputenc}
\usepackage[T1]{fontenc}
\usepackage{babel} 
\usepackage[vlined,ruled,linesnumbered]{algorithm2e}
\usepackage{algpseudocode}
\usepackage[toc,page]{appendix}
\usepackage{makecell}
\usepackage{xcolor,soul}
\definecolor{paleyellow}{HTML}{FFEF77}
\definecolor{paleorange}{HTML}{FBB068}
\definecolor{paleblue}{HTML}{65B2FF}
\definecolor{lemon}{HTML}{FDFFCC}

\DeclareRobustCommand{\hllemon}[1]{{\sethlcolor{lemon}\hl{#1}}}

\definecolor{Gray}{gray}{0.95}
\definecolor{LightCyan}{rgb}{0.88,1,1}

\newcommand{\code}[1]{{\ttfamily#1}}

\newcommand{\changed}[1]{{#1}}

\newcommand{\none}[1]{#1}

\newcommand{\fullmodel}{\textbf{Mix}ed \textbf{C}ontrastive \textbf{L}earning\xspace}
\newcommand{\model}{MixCL\xspace}
\newcommand{\kb}{KB-based\xspace}
\newcommand{\lm}{LM-based\xspace}

\title{Contrastive Learning Reduces Hallucination in Conversations}
\author {
Weiwei Sun\textsuperscript{\rm 1}, Zhengliang Shi\textsuperscript{\rm 1}, Shen Gao\textsuperscript{\rm 1}, Pengjie Ren\textsuperscript{\rm 1}, Maarten de Rijke\textsuperscript{\rm 2}, Zhaochun Ren\textsuperscript{\rm 1}\thanks{Corresponding author.}
}
\affiliations {
\textsuperscript{\rm 1}Shandong University, Qingdao, China\\
\textsuperscript{\rm 2}University of Amsterdam, Amsterdam, The Netherlands\\
\{weiwei.sun,shizhl\}@mail.sdu.edu.cn, \{shengao,renpengjie,zhaochun.ren\}@sdu.edu.cn, m.derijke@uva.nl
}

\usepackage{bibentry}

\begin{document}
\maketitle

\begin{abstract}

%
Pre-trained language models (LMs) store knowledge in their parameters and can generate informative responses when used in conversational systems.
However, LMs suffer from the problem of ``hallucination:'' they may generate plausible-looking statements that are irrelevant or factually incorrect.
To address this problem, we propose a contrastive learning scheme, named \model.
A novel mixed contrastive objective is proposed to explicitly optimize the implicit knowledge elicitation process of LMs, and thus reduce their hallucination in conversations.
We also examine negative sampling strategies of retrieved hard negatives and model-generated negatives.
We conduct experiments on Wizard-of-Wikipedia, a public, open-domain knowledge-grounded dialogue benchmark, and assess the effectiveness of \model.
\model effectively reduces the hallucination of LMs in conversations and achieves the highest performance among \lm dialogue agents in terms of relevancy and factuality.
We show that \model achieves comparable performance to state-of-the-art \kb approaches while enjoying notable advantages in terms of efficiency and scalability.
\end{abstract}


\section{Introduction}
\label{section:introduction}

Open-domain dialogue agents have received increasing attention in recent years~\citep{DeFreitas2020TowardsAH,Huang2020ChallengesIB}.
In an engaging open-domain dialogue, a large amount of knowledge, such as commonsense~\citep{Young2018AugmentingED} and factual knowledge~\citep{Dinan2019WizardOW}, is involved.
To integrate knowledge into dialogue agents, \emph{\kb methods} have been proposed to explicitly acquire knowledge from knowledge bases~\citep{Young2018AugmentingED,Dinan2019WizardOW}.
However, \kb methods suffer from problems of retrieval error~\citep{Liu2022MultiStagePF} and inefficiency~\citep{Xu2021RetrievalFreeKD}.
Meanwhile, recent years have witnessed a rapid development of pre-trained language models (LMs)~\citep{Devlin2019BERTPO,Brown2020LanguageMA} and their applications to dialogue tasks~\citep{Thoppilan2022LaMDALM}.
Large LMs implicitly store knowledge in their parameters during the pretraining stage~\citep{Petroni2019LanguageMA,Zhou2020EvaluatingCI} and thus, to some extent, they can serve as knowledge bases to ground open-domain dialogues~\citep{Zhao2020ArePL}.
Such approaches, known as \emph{\lm methods}, achieve promising performance in generating informative responses and obviate the drawbacks of \kb methods.
However, \lm methods have the problem of ``hallucination''~\citep{Shuster2021RetrievalAR,Ji2022SurveyOH}: they generate plausible-looking statements that are irrelevant or factually incorrect.

\begin{figure}[t]
 \centering
\includegraphics[width=1.0\columnwidth]{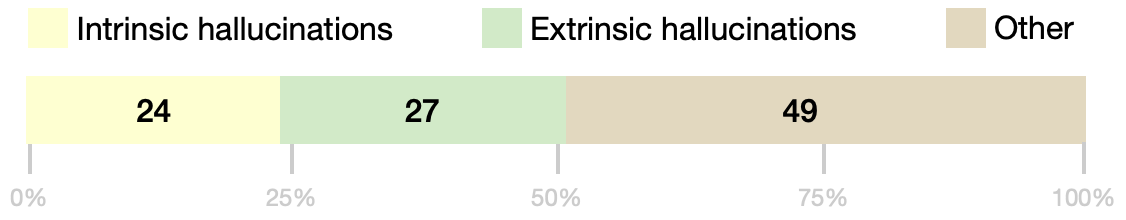} 
\caption{
Results of a pilot experiment where annotators were asked to label 200 responses generated by BART on the Wizard-of-Wikipedia dataset for hallucination.
}
\label{fig:motivation}
\end{figure}

To understand the severity of hallucinations of LMs, we conduct a pilot experiment.
We sample 200 responses generated by BART~\citep{Lewis2020BARTDS} on the Wizard-of-Wikipedia dataset~\citep{Dinan2019WizardOW} for various topics and conversation turns.
These responses are annotated by three well-informed experts in terms of knowledge relevancy and factuality. 
Based on the results, we group the hallucinations of LMs into two types: \emph{intrinsic hallucinations} and \emph{extrinsic hallucinations}.
Intrinsic hallucinations are non-factual statements, such as incorrectly predicting a celebrity's birthday.
Extrinsic hallucinations are irrelevant or out-of-context responses, such as the a description of the history of football when the user asks the number of teams  currently in the NFL. 
Fig.~\ref{fig:motivation} summarizes the outcomes: 
intrinsic and extrinsic hallucinations account for 24\% and 27\% of the responses, respectively.

The problem of hallucinations is mainly attributable to the optimization recipes: the commonly used maximum likelihood estimation (MLE) with teacher forcing training encourages the model to imitate the training data blindly, leading to model hallucinations at inference time~\citep{Kang2020ImprovedNL}.
Most studies on tackling hallucination in conversations focus on \kb methods and use pre-retrieval~\citep{Shuster2021RetrievalAR} or post-editing techniques~\citep{Dziri2021NeuralPH} to improve faithfulness; the hallucination of \lm agents in eliciting knowledge inside LMs' parameters is still underexplored.

In this paper, we propose \fullmodel (\model) to alleviate the hallucinations of \lm dialogue agents.
\model explicitly samples the most confusing knowledge to the model and reduces its generation probability by contrasting it with the ground-truth.
To this end, two novel steps are used by \model:
\begin{enumerate*}[label=(\roman*)]
\item negative sampling, and 
\item mixed-contrastive learning.
\end{enumerate*}
In the former, we sample the most confused negative knowledge by retrieving from the corpus or deriving via model bootstrapping.
In the latter, we propose mixed-contrastive learning under the inspiration of mix-up data augmentation~\citep{Zhang2018mixupBE}, which mixes the positive and negative at span level.
Moreover, we propose two mixed strategies regarding the two types of hallucination: entity-based mix-up and constituency-based mix-up.
Finally, \model is optimized in an end-to-end manner, thus avoiding the retrieval step during inference and instead using the knowledge inside its parameters.

We conduct experiments on Wizard-of-Wikipedia~\citep{Dinan2019WizardOW}, an open-domain, knowledge-grounded dialogue dataset.
Extensive experiments show that \model improves the informativeness and relevancy of the responses.
Compared with previous \lm methods~\citep{Zhao2020ArePL,Xu2021RetrievalFreeKD,Liu2022MultiStagePF}, \model achieves improvements by 5\% to 15\% in terms of response quality and relevancy.
Moreover, \model achieves comparable performance as state-of-the-art \kb methods (e.g.,  KnowledGPT~\citep{Zhao2020KnowledgeGroundedDG}), while speeding up 5$\times$ in model inference and showing superior scalability.
The effectiveness of \model is also verified through human evaluation and ablation experiments.

Our contributions are as follows:
\begin{enumerate*}[label=(\roman*)]
    \item We propose \model, which reduces hallucinations of LMs in conversation through contrastive learning.
    \item We propose a hard negative sampling strategy to obtain the most confused negative knowledge (see Section~\ref{sec:neg}).
    \item We propose a mix contrastive objective to optimize the model at span level (see Section~\ref{sec:cl}).
    \item Experiments on the Wizard-of-Wikipedia dataset show that \model effectively reduces the hallucinating content produced by the LM and achieves comparable performance to \kb approaches.\footnote{We release our code at \url{https://github.com/sunnweiwei/MixCL}.}
\end{enumerate*}


\section{Related work}

\subsection{Knowledge-grounded dialogues}
In open-domain knowledge-grounded dialogues (KGDs), people respond to each other’s utterances in a meaningful way by integrating knowledge~\cite{Young2018AugmentingED,Huang2020ChallengesIB}.
To integrate knowledge, \emph{\kb methods} have been explored~\citep{Liu2018KnowledgeDF,Young2018AugmentingED,Dinan2019WizardOW}; they retrieve knowledge from a corpus through additional information retrieval (IR) modules.
Studies on \kb methods focus on knowledge selection~\citep{Meng2020DukeNetAD,Shuster2021RetrievalAR} and knowledge-grounded response generation~\citep{Zhao2020KnowledgeGroundedDG,Zheng2021ExploringPF}.
However, \kb methods suffer from the problems of retrieval errors~\citep{Liu2022MultiStagePF}, inefficiencies~\citep{Xu2021RetrievalFreeKD}, and multi-granularity knowledge integration~\citep{Wu2022LexicalKI}.

\subsection{Language models as knowledge bases}
Recent years have witnessed a rapid development of language models (LMs)~\citep{Brown2020LanguageMA} and \lm dialogue agents~\citep{Thoppilan2022LaMDALM}.
Large LMs store knowledge into their parameters during pre-training and can generate informative responses in conversations~\citep{Zhao2020ArePL}.
\citet{Petroni2019LanguageMA} show that LMs can serve as knowledge bases for downstream tasks (e.g., question answering~\citep{Roberts2020HowMK}). 
On this basis, \citet{Zhao2020ArePL} show that LMs can ground open-domain dialogues using their implicit knowledge.
\citet{Madotto2020LearningKB} embed knowledge bases into model's parameters for end-to-end task-oriented dialogues.
\citet{Roller2021RecipesFB} finetune LMs on KGD data.
\citet{Cui2021KnowledgeEF} propose knowledge-enhanced finetuning methods to handle unseen entities.
\citet{Xu2021RetrievalFreeKD} propose a topic-aware adapter to adapt LMs in KGDs.
\citet{Liu2022MultiStagePF} propose a multi-stage prompting approach for triggering knowledge in LMs.
\citet{Wu2022LexicalKI} propose lexical knowledge internalization to integrate token-level knowledge into the model’s parameters.
However, existing \lm methods suffer from the problem of hallucination.
In this paper, we optimize the implicit knowledge eliciting process, i.e., reduce hallucination of LMs in KGD, via the proposed contrastive learning framework \model.

\subsection{Contrastive learning}
Contrastive learning (CL)~\citep{Chopra2005LearningAS,Chen2020ASF} is based on the idea that similar samples should also be close in representation space, and has seen applications in NLP~\citep{Gao2021SimCSESC}.
CL has been used for optimizing knowledge retrieval processes~\citep{Karpukhin2020DensePR,Xiong2021ApproximateNN}, where the model learns to identify positive knowledge from negatives.
On the task of neural text generation, CL~\citep{Jiang2022ASC}, a.k.a.\ unlikelihood training~\citep{Welleck2020NeuralTG} or negative training~\citep{He2020NegativeTF}, alleviates undesirable properties of the generated output, e.g., repetition~\citep{Shirai2020NeuralTG,Jiang2022ASC}, maliciousness~\citep{He2020NegativeTF}, dullness~\citep{Li2020EnhancingDG,Li2022DiversifyingND}, or inconsistency~\citep{Li2020DontST}.
Moreover, \citet{Cao2021CLIFFCL} propose a sentence level contrastive learning method to reduce the hallucinations of text summarization model.
Unlike existing studies, we propose a mixed contrastive learning framework \model that eliminates the hallucination at the span level with effective negative sampling strategies.


\section{Problem formulation}
\begin{figure}[t]
 \centering
 \begin{subfigure}{0.49\columnwidth}
 \includegraphics[width=1.0\textwidth]{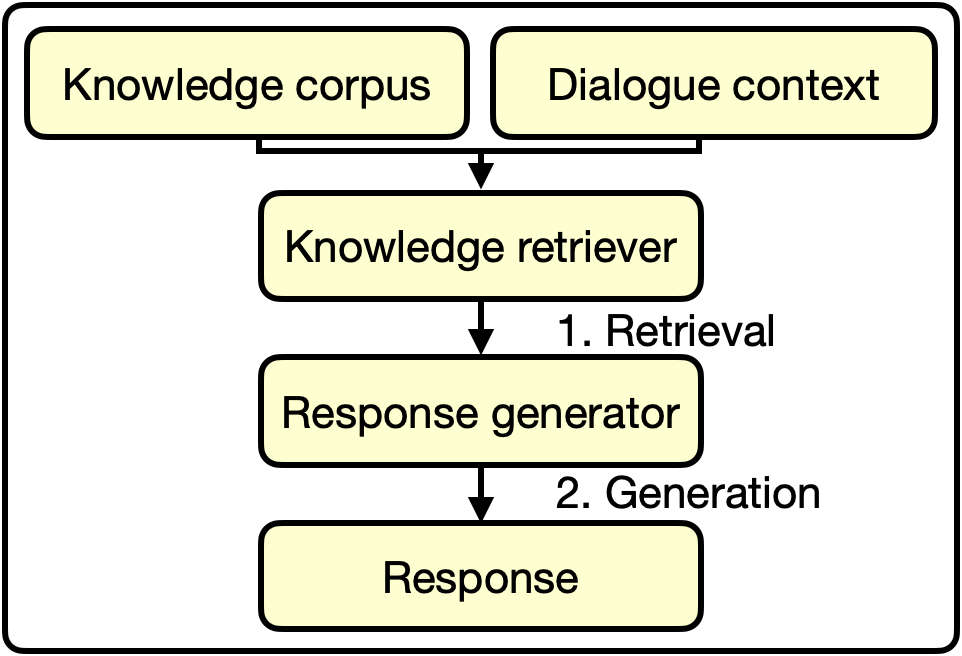}
 \caption{\kb dialogue agents explicitly retrieve text-based knowledge from corpus.\\ }
 \end{subfigure}
 \hfill
 \begin{subfigure}{0.49\columnwidth}
 \includegraphics[width=1.0\textwidth]{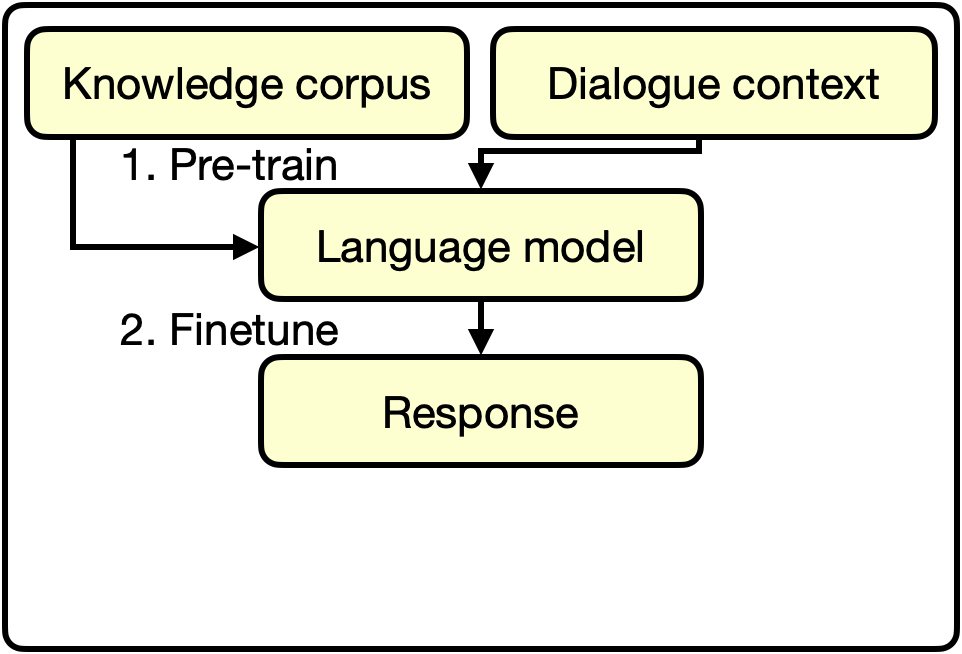}
 \caption{\lm dialogue agents store knowledge in LM para\-meters and generate responses using implicit knowledge.}
 \end{subfigure}
\vspace*{-2mm}
 \caption{Types of dialogue agents.}
\label{fig:lm-as-kb}
\end{figure}

Let $x$, $y$, and $k$ be the dialogue context, the corresponding response, and the ground-truth knowledge, respectively.
As illustrated in Fig.~\ref{fig:lm-as-kb}, given a knowledge corpus $\mathcal{K}$, a dialogue agent learns to predict an informative response $y$ based on the dialogue context $x$ using the knowledge in $\mathcal{K}$.
As discussed earlier, two approaches are studied in KGD, \emph{\kb methods} and \emph{\lm methods}. 
In this paper, we focus on the latter one.

\paragraph{\kb methods.}
\kb dialogue agents~\citep{Dinan2019WizardOW} ground the response generation by explicitly retrieving knowledge from $\mathcal{K}$.
Two sub-modules, i.e., knowledge retriever and response generator, are employed by \kb approaches, as shown in Fig.~\ref{fig:lm-as-kb} (a).

\paragraph{\lm methods.}
In this paper, we explore language models as knowledge bases for dialogue agents~\citep{Zhao2020ArePL,Xu2021RetrievalFreeKD}, as illustrated in Fig.~\ref{fig:lm-as-kb} (b).
In \lm approaches, the LMs are first pre-trained on $\mathcal{K}$ to store the knowledge in their parameters.
Then, the models directly generate $y$ given $x$ using the knowledge in their parameters and getting rid of the explicit retrieval step.

\section{Preliminaries}
We propose a \lm dialogue agent for open-domain KGD.
The proposed model $p_\theta(y|x)$ is based on a transformer-based language model with encoder-decoder architecture.
The model is first pre-trained on the corpus $\mathcal{K}$ and then finetuned on dialogue data to generate informative responses.

\paragraph{Pre-training on knowledge corpus.}
We employ BART \citep{Lewis2020BARTDS} as the pre-trained transformer, which is pre-trained by denoising self-supervised learning:
\begin{equation} \label{eq:lm}
    \mathcal{L}_{\text{LM}} = -\mathbb{E}_{k \sim \mathcal{K}}~\log p_\theta(k|\hat{k}),
\end{equation}
where \changed{$\mathcal{K}$ is a text-based knowledge corpus (e.g., Wikipedia),} $k$ is a text sampled from knowledge corpus $\mathcal{K}$, and \smash{$\hat{k}$} denotes corrupted text by corruption functions (e.g., masking, deletion, infilling, etc.; \citet{Lewis2020BARTDS}).

\paragraph{Finetuning on dialogue datasets.}
With the pre-trained LM, the model generates the response $y$ given $x$ without explicit knowledge retrieval step~\citep{Zhao2020ArePL,Xu2021RetrievalFreeKD}.
Maximum likelihood estimation (MLE) training loss on dialogue data with paired $(x, y)$ is employed by previous methods.
In MLE, the model learns to predict the ground-truth tokens for each step in a teacher forcing paradigm~\citep{Zhao2020ArePL,Xu2021RetrievalFreeKD}:
\begin{equation} \label{eq:mle}
    \mathcal{L}_{\text{MLE}} = -\log p_\theta(y|x) = -\sum_{t=1}^{|y|} \log p_\theta(y_t|y_{<t},x).
\end{equation}
However, despite its effectiveness in generating informative responses, MLE loss encourages the model to imitate the training data blindly and leads to model hallucination~\citep{Kang2020ImprovedNL}.
Studies have found that models trained with standard MLE may over-rely on previously predicted tokens, exacerbating error propagation~\citep{Wang2020OnEB}.
As a result, during the inference stage, as the generated sequence grows, the errors accumulate along the sequence, and the model tends to amplify errors and generate hallucinating contents. 
We propose a novel contrastive learning framework \model to address this problem.

\begin{figure*}[!t]
 \centering
 \includegraphics[width=2\columnwidth]{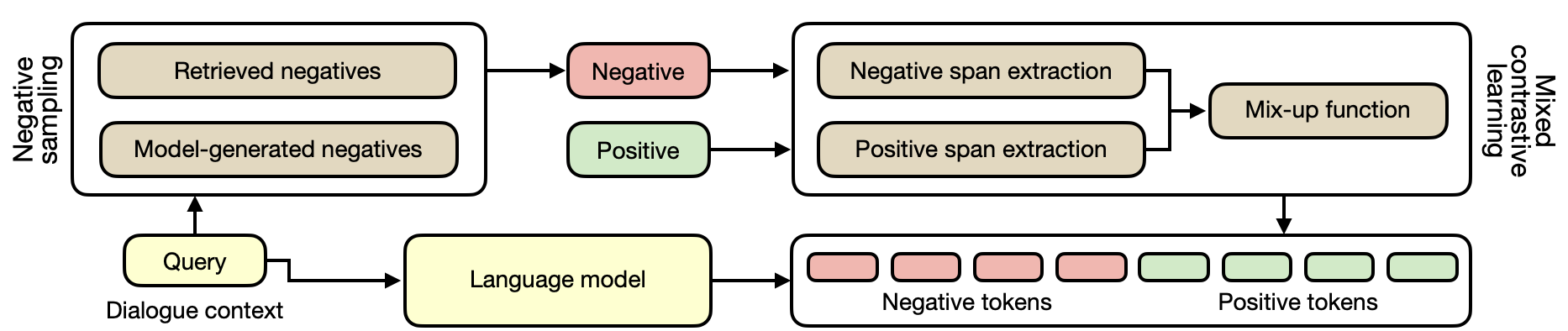}
 \caption{Overview of \model. \model consists of two steps: (i) negative sampling (Section~\ref{sec:neg}), which samples most confusing negative knowledge to the model, and (ii) mixed contrastive learning (Section \ref{sec:cl}), which reduces the generation probability of negative tokens through mixed contrastive learning.}
 \label{fig:model}
\end{figure*}

\section{\model}
Next, we present the proposed \model framework for addressing the hallucination of LMs.
\model explicitly samples negative knowledge (i.e., non-factual or irrelevant knowledge) and reduces the generation probability of negative tokens by LMs through contrastive learning.
As illustrated in Fig.~\ref{fig:model}, \model consists of two steps: \emph{negative sampling} and \emph{mixed contrastive learning}.
In this section, we first present the negative sampling methods, then the mixed contrastive learning, and finally our optimization strategies.

\subsection{Negatives sampling} 
\label{sec:neg}
We sample negative knowledge for the dialogue context to construct training examples for contrastive learning.
Formally, let $z^+$ be positive knowledge, i.e., a factual and relevant knowledge snippet, and let $\mathcal{Q}_{\text{Pos}}(x)$ be the collection of positive knowledge regarding $x$, where the $z^+ \sim \mathcal{Q}_{\text{Pos}}(x)$ is sampled from it.
Here, $\mathcal{Q}_{\text{Pos}}(x)$ can be obtained through human labeling~\citep{Dinan2019WizardOW} or heuristic methods~\citep{Zhao2020KnowledgeGroundedDG}.
We define $z^-$ as negative knowledge, i.e., a non-factual or irrelevant knowledge snippet for $x$.
Then, negative sampling is applied to construct the snippets $z^-$ where the model is most likely to get confused.
We introduce two methods for negative sampling, i.e., \emph{retrieved negatives} and \emph{model-generated negatives}, as illustrated in Fig.~\ref{fig:model}.

\paragraph{Retrieved negatives.}
For a given $x$, a retrieval tool $\text{Ret}(*)$ is employed to retrieve irrelevant but potentially confusing knowledge from knowledge corpus $\mathcal{K}$:
\begin{equation} \label{eq:ret}
    \mathcal{Q}_{\text{Ret}}(x) = \{z^-|z^-\in\text{Ret}(x, \mathcal{K}), z^-\notin\mathcal{Q}_{\text{Pos}}(x)\},
\end{equation}
where $\text{Ret}(\cdot,\cdot)$ is implemented as TF-IDF retriever~\citep{Dinan2019WizardOW}, and $z^-\notin\mathcal{Q}_{\text{Pos}}(x)$ imposes the constraint that negative knowledge snippets should not be included in the positive knowledge.

\paragraph{Model-generated negatives.}
We also exploit a model bootstrapping approach, in which we generate knowledge by a model $p_\theta(z|x)$ and retain the examples where hallucination exist.
We define:
\begin{equation} \label{eq:model}
    \mathcal{Q}_{\text{Model}}(x) = \{z^-|z^-\sim p_\theta(z|x), z^-\cap\mathcal{Q}_{\text{Pos}}(x)=\varnothing\},
\end{equation}
where $z^-\sim p_\theta(z|x)$ denotes a negative knowledge snippet sampled from the LM with $\theta$, and $z^-\cup\mathcal{Q}_{\text{Pos}}(x)=\varnothing$ imposes the constraint that negative knowledge snippets should not be included in the positive knowledge, which is implemented with a natural language inference (NLI) toolkit.\footnote{\url{https://huggingface.co/roberta-large-mnli}}


On the basis of the above two methods, we define the constructed negative collection $\mathcal{Q}_{\text{Neg}}(x)$ with a hyper-parameter $\beta \in [0,1]$ to control the relative contribution of the methods:
\begin{equation} \label{eq:neg}
    \mathcal{Q}_{\text{Neg}}(x) = \beta \mathcal{Q}_{\text{Ret}}(x) + (1 - \beta) \mathcal{Q}_{\text{Model}}(x).
\end{equation}


\subsection{Mixed contrastive learning} \label{sec:cl}
Based on the positive knowledge $z^+$ and the sampled negative knowledge $z^-$, we introduce a contrastive learning framework to identify positive knowledge from negatives:
\begin{equation} \label{eq:cl}
    \mathcal{L}_{\text{CL}} = \mathbb{E}_{z^+\sim\mathcal{Q}_{\text{Pos}}(x),\{z^-_i\}_{i=1}^{M}\overset{\text{iid}}{\sim}\mathcal{Z}} l(x,z^+,\{z^-_i\}_{i=1}^{M},\theta),
\end{equation}
where $l$ denotes a contrastive loss function that is typically defined as cross-entropy loss $l_{\text{ce}}$\footnote{$l_{\text{ce}}(x,z^+\!,\{z^-_i\}_{i=1}^{M},\theta)\!=\!-\log \frac{\exp{p_\theta(z^+|x)}}{\exp{p_\theta(z^+|x)}+\sum_{i=1}^M\exp{p_\theta(z^-_i|x)}}$.}~\citep{Gao2021SimCSESC,Cao2021CLIFFCL}, and $M$ denotes the number of negatives.

However, $l_{\text{ce}}$ only considers token-level or sentence-level contrast. It ignores fine-grained \emph{span-level} contrast even though hallucinations typically exists at the span level. 
Therefore, inspired by work on mix-up data augmentation~\citep{Zhang2018mixupBE,Kim2020MixCoMC,Shi2021SubstructureSS,Zhang2022TreeMixCC}, we propose a \emph{mixed contrast objective}, which mixes the positive and negative examples into a sequence at the span level.
As illustrated in Fig.~\ref{fig:model}, the proposed mixed contrastive learning method has three parts: 
\begin{enumerate*}[label=(\roman*)]
\item \emph{extracting spans}, which extracts meaningful spans from both positive and negative knowledge; 
\item \emph{mixing examples}, which mixes positive and negative knowledge using the extracted spans; and 
\item \emph{mixed-contrast loss}, which optimizes the model at the span level through contrastive learning.
\end{enumerate*}

\subsubsection{Extracting spans.} \label{sec:extraction}
We extract the key components from the both positive and negative knowledge, $z^+$ and $z^-$.
Regarding the two types of hallucinations, i.e., the intrinsic and extrinsic, we design two extraction strategies.
As part of the pilot experiment reported in Section~\ref{section:introduction}, we find that intrinsic hallucinations are typically associated with confused entities.
Therefore, we use \emph{named entity recognition} (NER)\footnote{\url{https://spacy.io/api/entityrecognizer/}} to extract entities of various types e.g. \code{person} and \code{time}.
Moreover, we find that extrinsic hallucination is mainly triggered by the emergence of irrelevant sentence fragments in the text.
Therefore, we use \emph{constituency parsing} (CP)\footnote{\url{https://stanfordnlp.github.io/stanza/constituency.html}} to extract sentence constituents, e.g., \code{noun} and \code{particle}.
Through the two strategies, we extract sequence spans from $z^+$ and $z^-$, respectively.

\vspace*{3mm}
\textbf{Example.}
Consider knowledge snippets about the French soccer player Thierry Henry.  A statement like \emph{He was born and raised in Paris} would be in $z^+$, while the span ``Montreal, Quebec, Canada'' could be extracted from a snippet such as in \emph{He was born in Montreal, Quebec, Canada} in $z^-$.

\subsubsection{Mixing examples.} 
\label{sec:mix-up}
Based on the extracted spans, we mix the two examples $z^+$ and $z^-$ into a mixed sequence $\tilde{z}$ via a mix-up function: $\tilde{z} = \text{Mix}(z^+,z^-)$.
The mix-up function randomly selects a span in $z^+$, and then selects a span with the same type in $z^-$ to substitute it.
We define a sequence $\phi$ with the same length of $\tilde{z}$, which annotates the tokens in $\tilde{z}$ as $1$ if they come from $z^+$ and $0$ if they come from $z^-$. 

In the earlier Thierry Henry example, the span ``Paris'' in a snippet in $z^+$  can be selected and substituted by the corresponding ones from a snippet in $z^-$, such as ``Montreal, Quebec, Canada.''

\subsubsection{Mixed-contrast loss.} \label{sec:loss}
Based on the mixed sequence $\tilde{z}$ and $\phi$, we design a loss function $l_{mix}$ as follows:
\begin{equation}
\begin{split} 
    l_{mix}(z^+,z^-)={}&\\
   \textstyle- \sum_{j=1}^{|\tilde{z}_i|} \lbrack~&\phi_{i,j}\log p_\theta(\tilde{z}_{i,j}|\tilde{z}_{i,<j},x) +{}\\ 
    &(1-\phi_{i,j}) \log (1 - p_\theta(\tilde{z}_{i,j}|\tilde{z}_{i,<j},x))\rbrack,
\end{split}
\label{eq:mix-loss}
\end{equation}
where $\tilde{z}_i=\text{Mix}(z^+,z^-_i)$ is a mixed sequence of $z^+$ and $z^-_i$, and $\phi_{i,j}$ denotes the sign of token $\tilde{z}_{i,j}$, which equals $1$ for positive tokens and $0$ for negative tokens.
Using the negative collection $\mathcal{Q}_{\text{Neg}}(x)$ defined in Eq.~\ref{eq:neg}, we formalize the mixed contrast objective $\mathcal{L}_{\text{MCL}}$ as:
\begin{equation}
\textstyle 
    {\sum}_{z^+\sim\mathcal{Q}_{\text{Pos}}(x)}{\sum}_{z^-_i\sim\mathcal{Q}_{\text{Neg}}(x)}^{i=1,...,M}
    l_{mix}(x,z^+,z^-_i,\theta).
\label{eq:mcl}
\end{equation}

\subsection{Optimization} \label{sec:optimization}
During finetuning, \model is optimized by minimizing $\mathcal{L}_{\text{MCL}}$.
Two additional loss are considered in training, i.e., $\mathcal{L}_{\text{LM}}$, $\mathcal{L}_{\text{MLE}}$.
$\mathcal{L}_{\text{LM}}$ is used to alleviate catastrophic knowledge forgetting~\citep{Devlin2019BERTPO,Chen2020RecallAL} and $\mathcal{L}_{\text{MLE}}$ is used to optimize the response generation ability. 
Therefore, the final training objective is defined as:
\begin{equation} \label{eq:all}
    \mathcal{J}(\theta) = \alpha_1 \mathcal{L}_{\text{MLE}} + \alpha_2 \mathcal{L}_{\text{MCL}} + \alpha_3 \mathcal{L}_{\text{LM}},
\end{equation}
where three losses are optimized jointly and $\alpha_1$, $\alpha_2$, $\alpha_3$ denote the weights of the three losses, respectively.


\section{Experimental setup}

\subsection{Datasets and evaluation metrics}
We conduct experiments on the Wizard of Wikipedia (WoW) dataset. 
WoW is built with crowd-sourcing and employs Wikipedia as the knowledge corpus. 
WoW consists of 22,311 conversations over 1,365 general topics that range from e-books to toga parties to showers.
The ground-truth knowledge used in each turn is manually labeled.
The WoW test set is split into \emph{test seen} and \emph{test unseen} based on whether the topic appears in the training set. 
We evaluate our methods on both test seen and test unseen.

We choose F1, ROUGE, BLEU, MT, Knowledge-F1 (KF1), Entity-F1 (EF1), and Accuracy (Acc) as metrics. 
\textbf{F1}~\citep{Dinan2019WizardOW} calculates the unigram F1 between the generated text and the ground-truth text.
For \textbf{ROUGE}~\citep{Lin2004ROUGEAP} we use ROUGE-L (RL for short) following previous work.
\textbf{BLEU}~\citep{Papineni2002BleuAM} we use BLEU-2 and BLEU-4 (or B2 and B4 for short) and use the implementation in the NLTK Toolkit.
\textbf{MT} (Meteor)~\citep{Denkowski2014MeteorUL} is based on the harmonic mean of unigram precision and recall. 
\textbf{Knowledge-F1}~\citep{Dinan2019WizardOW} (or KF1 for short) calculates the F1 between the generated response and the ground-truth knowledge sentence, which indicates the informativeness of a response.
\textbf{Acc} measures the knowledge selection accuracy. As we skip the knowledge selection step, we select knowledge by matching the generated response with each knowledge candidate in WoW using the F1 score.
\textbf{Entity-F1} (or EF1 for short) identifies entities in text using Spacy, deletes the non-entity words, and calculates the F1 score between the modified generated text and the ground-truth response. EF1 eliminates the impact of the stop-word and focuses on the accuracy of entities.

In addition, we randomly sample 100 examples from the test seen and test unseen segments of the test set, respectively, and recruit three experts for human evaluation.
Each annotator is presented with examples that come with dialogue context and model responses. 
Four metrics are considered in the human evaluation:
\textbf{Informativeness}, which measures whether the response is knowledge-inclusive;
\textbf{Relevancy}, which measures whether the response's content is relevant to the dialogue;
\textbf{Factuality}, which measures whether the information in the response is factually correct;\footnote{\changed{The human annotators used Google to check the factuality of the responses.}} and
\textbf{Humanlikeness}, which measures whether the response is human-like in its fluency and naturalness.
The annotators are asked to assign a score in $\{0, 1\}$ (representing ``non-factual'' and ``factual'') for factuality, and a score in $\{0, 1, 2\}$ (representing ``bad'' ``fair'', and ``good'') for the others.

\begin{table*}[!t]
\centering
\small
\setlength\tabcolsep{4.5pt}
\caption{Evaluation results on Wizard-of-Wikipedia.
The first group lists \emph{\kb methods under realistic conditions}. The second group lists \emph{\kb methods under oracle conditions}.
The third group lists \emph{\lm methods}, including \model.
We highlight the results of \model that significantly exceed the previous-best \lm methods in \textbf{boldface} (t-test, $p<0.05$), and of \model that exceed the best \kb methods under realistic conditions in \hllemon{yellow}.
We also highlight the best results of previous \kb methods and \lm methods by \underline{underlining} them, respectively.
}
\label{table:main}

\begin{tabular}{@{}l cccccccc cccccccc @{}}

\toprule
& \multicolumn{8}{c}{\textbf{Test seen}} 
& \multicolumn{8}{c}{\textbf{Test unseen}} 
\\
\cmidrule(lr){2-9} \cmidrule(lr){10-17} 
\textbf{Method}
& F1 & RL & B2 & B4 & MT & KF1 & EF1 & Acc & F1 & RL & B2 & B4 & MT & KF1 & EF1 & Acc\\

\midrule
\multicolumn{4}{@{}l}{\emph{\kb methods under realistic conditions}}\\

TMN~\citep{Dinan2019WizardOW}
& 17.3 & 17.0 & 5.7 & 1.1 & 14.8 & 15.8 & \phantom{0}8.7 & 15.2 
& 14.4 & 14.5 & 3.3 & 0.3 & 11.5 & \phantom{0}9.4 & \phantom{0}2.1 & \phantom{0}8.6 \\

DukeNet~\citep{Meng2020DukeNetAD}
& 18.5 & 17.7 & 6.4 & 1.9 & 16.0 & 18.5 & 12.0 & 20.6 
& 15.9 & 15.9 & 4.8 & 1.1 & 13.7 & 14.7 & \phantom{0}8.0 & 14.3 \\

KnowledGPT~\citep{Zhao2020KnowledgeGroundedDG}
& \underline{21.1} & \underline{20.1} & \underline{8.9} & \underline{3.4} & \underline{20.0} & \underline{22.2} & \none{15.5} & \underline{24.3}
& 19.5 & \underline{18.4} & 8.0 & 2.6 & \underline{18.3} & 20.0 & 11.7 & 20.2\\

KnowBART
& \none{21.1} & 18.9 & 8.5 & 3.3 & 17.8 & 21.3 & \underline{16.2} & 24.2 
& \underline{21.0} & 18.3 & \underline{8.9} & \underline{3.6} & 17.9 & \underline{22.5} & \underline{16.2} & \underline{24.0}\\

\midrule
\multicolumn{4}{@{}l}{\emph{\kb methods under oracle conditions}}\\
DukeNet~\citep{Meng2020DukeNetAD}
& 19.3 & 18.7 & 7.5 & 2.5 & 17.2 & 19.6 & 13.2 & 22.1 
& 17.1 & 17.0 & 6.0 & 1.7 & 15.2 & 16.5 & \phantom{0}9.2 & 16.8 \\


KnowledGPT~\citep{Zhao2020KnowledgeGroundedDG}
& 22.0 & 20.8 & 9.9 & 3.7 & 20.9 & 23.8 & 16.9 & 26.3 
& 20.5 & 19.5 & 8.7 & 3.0 & 19.3 & 22.1 & 13.3 & 22.6 \\

\changed{KnowBART}
& 22.1 & 19.6 & 9.1 & 3.7 & 18.1 & 23.1 & 18.0 & 26.8
& 22.7 & 20.1 & 9.8 & 4.3 & 18.7 & 24.1 & 18.4 & 27.5
\\

\midrule
\emph{\lm methods}\\
GPT-2~\citep{Zhao2020ArePL} 
& \underline{19.6} & 18.5 & \underline{7.8} & 1.4 & 17.8 & 17.9 & 13.3 & 15.4
& \underline{18.3} & \underline{17.3} & \underline{6.5} & 0.8 & 16.1 & 14.6 & 7.2 & \phantom{0}8.4 \\



BlenderBot~\citep{Roller2021RecipesFB} 
& 18.8 & \underline{19.4} & 7.7 & \underline{2.3} & 18.0 & 18.2 & 13.1 &  16.7
& 17.8 & 16.9 & 5.5 & 0.8 & 15.0 & 15.7 & 7.1 & \phantom{0}9.6 \\

KnowExpert~\citep{Xu2021RetrievalFreeKD} 
& 18.7 & 18.6 & 6.7 & 1.3 & 16.5 & 14.1 & \phantom{0}9.8 & 12.6 
& 16.7 & 17.2 & 5.4 & 0.6 & 14.5 & 11.8 & 5.5 & \phantom{0}9.2 \\

MSDP~\citep{Liu2022MultiStagePF} 
& 17.8 & 16.5 & 6.1 & 1.9 & \underline{18.2} & \underline{21.7} & \underline{13.9} & \underline{18.4}
& 16.9 & 16.1 & 5.5 & \underline{1.1} & \underline{16.2} & \underline{20.3} & \underline{8.4} & \underline{16.1} \\
\midrule

\textbf{Ours} 
& \hllemon{\textbf{21.6}} & \hllemon{\textbf{20.5}} & \hllemon{\textbf{9.2}} & \textbf{2.7} & \hllemon{\textbf{20.5}} &  \hllemon{\textbf{22.3}} & \hllemon{\textbf{16.3}} & \textbf{20.4} 
& \textbf{19.6} & \hllemon{\textbf{18.8}} & \textbf{7.4} & \textbf{1.4} & \textbf{18.0} & \none{18.0} & \textbf{11.6} & \none{14.4} \\


\bottomrule
\end{tabular}

\end{table*}

\subsection{Baselines}
We compare \model with baselines of two categories: 
\begin{enumerate*}[label=(\roman*)]
\item \emph{\kb methods} that use additional IR modules for explicit knowledge retrieval, and
\item \emph{\lm methods} that use LMs as a knowledge base.
\end{enumerate*}
All models are re-evaluated with the same evaluation function using the official public checkpoints.

The \kb methods we consider are:
\textbf{TMN}~\citep{Dinan2019WizardOW} (50M), which combines a transformer with an external memory network to select knowledge and generate a response;
\textbf{DukeNet}~\citep{Meng2020DukeNetAD} (150M), which is the best performing \kb method without using pre-trained LMs and which models knowledge shift with a dual learning scheme;
\textbf{KnowledGPT}~\citep{Zhao2020KnowledgeGroundedDG} (227M), which exploits pre-trained LMs in a \kb approach, selects knowledge using BERT, generates responses using GPT-2, and optimizes the two modules jointly with reinforcement learning; it achieves state-of-the-art performance. 
We also introduce \textbf{KnowBART} (600M), a \kb model that selects knowledge using RoBERTa and generates responses using BART-Large.

The \kb methods listed above retrieve knowledge under \emph{oracle conditions}, i.e., they are given a small subset of Wikipedia with roughly ten passages that \emph{definitely} contain the ground-truth knowledge~\citep{Dinan2019WizardOW,Liu2022MultiStagePF}.
We also consider \kb methods under \emph{realistic experimental conditions}, where passages from the full knowledge corpus (i.e., Wikipedia) are retrieved.
We employ the state-of-the-art passage retrieval model GENRE~\citep{DeCao2021AutoregressiveER}
from the KILT leaderboard~\citep{Petroni2021KILTAB},
which is reported to outperform competitors (e.g., DPR and BM25) by a substantial margin on WoW.

The \lm methods that we consider are:
\textbf{GPT-2}~\citep{Zhao2020ArePL} (345M),which finetunes GPT-2 on knowledge-grounded dialogue data;
\textbf{BlenderBot}~\citep{Roller2021RecipesFB} (400M), which pre-trains a transformer with encoder-decoder architecture on \emph{reddit} data, and then finetunes the model on KGD data;
\textbf{KnowExpert}~\citep{Xu2021RetrievalFreeKD} (117M), which uses a topic-aware adapter that first clusters Wikipedia using a topic model and then employs a mix-of-adapter architecture to adapt a GPT-2 model to open-domain dialogues;
\textbf{MSDP}~\citep{Liu2022MultiStagePF} (357M), which uses a multi-stage prompting model, designs task-specific prompts with task instructions and in-context examples, and uses Megatron-LM~\citep{Shoeybi2019MegatronLMTM} to produce knowledge and response in a two-stage process.

\subsection{Implementation details}
We implement \model using BART-Large (400M)~\citep{Lewis2020BARTDS} in HuggingFace’s Transformers library.
We use Wikipedia as the knowledge corpus $\mathcal{K}$, as it is used as knowledge corpus by WoW. 
We determine the hyperparameters through pilot experiments.
We set the weight of the language model loss $\alpha_3$ to $0.3$ at initialization and linearly decay until $0$.
We set $\alpha_1$ and $\alpha_2$, i.e., the weight of the MLE loss and MCL loss, to $0.4$ and $0.3$, respectively, and linearly increase to $0.5$ and $0.5$.
\changed{We use greedy decoding in testing.}
More details are available in Appendix~\ref{sec:implementation} or at \url{ https://github.com/sunnweiwei/MixCL}.


\section{Experimental results}

\subsection{Results of automatic evaluation}
Table \ref{table:main} shows the results of automatic evaluation
metrics. Overall, \model achieves the highest scores of the \lm methods and competitive results compared to the \kb methods under realistic conditions.
Compared with previous \lm methods (the third group in Table~\ref{table:main}), \model achieves the highest scores on almost all metrics.
For example, \model gets F1${}={}$21.6, B4${}={}$2.7 on test seen and F1${}={}$19.6, B4${}={}$1.4 on test unseen, with about 5\% to 15\% relative improvements over previous-best \lm baselines.
Moreover, we find a dilemma with the previous \lm methods in terms of response quality (e.g., F1, RL, B2) and knowledge relevance (e.g., KF1, EF1, Acc).
For example, MSDP performs well on knowledge relevance at the expense of response quality, while GPT-2 and BlenderBot show the opposite.
\model, on the other hand, performs well on both fronts.

Furthermore, compared with \kb methods (the first block in Table~\ref{table:main}), we find that \model outperforms two non-LM methods (DukeNet and TMN) by a large margin.
Compared to KnowledGPT and KnowBART, which combine LMs with the \kb approach, \model outperforms them on test seen.
On test unseen, \model lags behind the best performing \kb baselines, probably due to knowledge forgetting issues.

Finally, under oracle conditions, the \kb methods (the second group in Table~\ref{table:main}) show better results than \model.
However, the manually selected knowledge candidates include the ground-truth, which is unavailable in realistic scenarios.

\begin{table}[!t]
\centering
\small
\setlength\tabcolsep{2.7pt}
\caption{Human evaluation results. Methods marked with $^\text{K}$ denote \kb methods, and those marked with $^\text{L}$ denote \lm methods. The four metrics (Info., Rel., Fact., and Hum.) denote informativeness, relevance, factuality, and humanlikeness, respectively.}
\label{table:human}

\begin{tabular}{@{}l cccc cccc@{}}

\toprule
\multirow{2}{*}{\textbf{Methods}}
& \multicolumn{4}{c}{\textbf{Test seen}} 
& \multicolumn{4}{c}{\textbf{Test unseen}}
\\
\cmidrule(r){2-5} \cmidrule{6-9}

{} & Info. & Rel.  & Fact.  & Hum. 
& Info. & Rel.  & Fact.  & Hum. \\

\midrule

DukeNet$^\text{K}$
& 1.44 & 1.22 & 0.71 & 1.16
& 1.21 & 1.08 & 0.72 & 1.03
\\

KnowledGPT$^\text{K}$
& 1.67 & 1.47 & 0.87 & 1.73
& 1.63 & 1.23 & 0.83 & 1.36
\\

KnowBART$^\text{K}$
& 1.67 & 1.57 & 0.89 & 1.70
& 1.68 & 1.56 & 0.91 & 1.44
\\

\midrule

KnowExpert$^\text{L}$
& 1.45 & 1.36 & 0.62 & 1.45
& 1.49 & 1.26 & 0.59 & 1.15
\\

MSDP$^\text{L}$
& 1.20 & 0.96 & 0.71 & 0.98
& 1.28 & 1.18 & 0.82 & 1.05
\\

BART$^\text{L}$
& 1.51 & 1.45 & 0.76 & 1.58
& 1.50 & 1.47 & 0.82 & 1.40
\\
\midrule
\textbf{Ours}$^\text{L}$
& 1.71 & 1.55 & 0.89 & 1.77
& 1.67 & 1.53 & 0.87 & 1.47
\\
\midrule
Human 
& 1.84 & 1.85 & 0.98 & 1.96
& 1.83 & 1.85 & 0.95 & 1.95
\\

\bottomrule
\end{tabular}
\end{table}

\subsection{Results of human evaluation}
Table~\ref{table:human} shows the human evaluation results. 
The Fleiss' kappa value is above $0.60$, indicating substantial agreement among the annotators.
\model consistently outperforms \lm baselines on all metrics, and also outperforms \kb baselines in metrics. 
\model is capable of generating more informative responses compared to previous \lm methods.
Moreover, \model effectively increases relevance and factuality, demonstrating its effectiveness in reducing both types of hallucinations.
In particular, we find that KnowledGPT is outperformed by \model in terms of knowledge relevance, probably due to the presence of retrieval errors.
Finally, \model's responses are considered more human-like by the annotators.

\begin{table}[!t]
\centering\small
\setlength\tabcolsep{1pt}
\caption{Ablation study. The base model, \model, is compared with several variants. See Section \ref{sec:ablation}.} 
\label{table:ablation}

\begin{tabular}{@{}l lll lll@{}}

\toprule
\multirow{2}{*}{\textbf{Methods}}
& \multicolumn{3}{c}{\textbf{Test seen}}
& \multicolumn{3}{c}{\textbf{Test unseen}}
\\
\cmidrule(lr){2-4} \cmidrule(lr){5-7}

{} & \multicolumn{1}{c}{F1} & \multicolumn{1}{c}{B4} & \multicolumn{1}{c}{KF1} 
& \multicolumn{1}{c}{F1} & \multicolumn{1}{c}{B4} & \multicolumn{1}{c}{KF1} \\

\midrule 
Base model
& 21.6 & 2.7 & 22.3
& 19.6 & 1.4 & 18.0
\\
\midrule

{-w/o $\mathcal{L}_{\text{MCL}}$}
& 21.0$_{\downarrow0.6}$ & 2.0$_{\downarrow0.7}$ & 19.1$_{\downarrow3.2}$
& 19.1$_{\downarrow0.5}$ & 1.0$_{\downarrow0.4}$ & 16.9$_{\downarrow1.1}$
\\

{-w/o $\mathcal{Q}_{\text{Neg}}(*)$} 
& 20.8$_{\downarrow0.8}$ & 2.4$_{\downarrow0.3}$ & 20.8$_{\downarrow1.5}$
& 19.0$_{\downarrow0.6}$ & 1.1$_{\downarrow0.3}$ &17.4$_{\downarrow0.6}$
\\

{-w/o $\mathcal{Q}_{\text{Model}}(*)$}~~ 
& 21.3$_{\downarrow0.3}$ & 2.5$_{\downarrow0.2}$ & 21.7$_{\downarrow0.6}$
& 19.4$_{\downarrow0.2}$ & 1.2$_{\downarrow0.2}$ &17.5$_{\downarrow0.5}$
\\

{-w/o $\mathcal{L}_{\text{LM}}$}
& 21.3$_{\downarrow0.3}$ & 2.6$_{\downarrow0.1}$ & 21.8$_{\downarrow0.5}$
& 18.6$_{\downarrow1.0}$ & 1.2$_{\downarrow0.2}$ & 16.7$_{\downarrow1.3}$
\\

{-Only $\mathcal{L}_{\text{MLE}}$} & 20.9$_{\downarrow0.7}$ & 1.8$_{\downarrow0.9}$ & 18.9$_{\downarrow3.4}$
& 18.8$_{\downarrow0.8}$ & 0.9$_{\downarrow0.5}$ & 16.0$_{\downarrow2.0}$
\\

\bottomrule
\end{tabular}
\end{table}

\subsection{Ablation studies} \label{sec:ablation}
In Table \ref{table:ablation}, we compare \model with several ablative variants.
The variants and our findings are as follows:

\noindent\textbf{No $\mathcal{L}_{\text{MCL}}$} -- 
We remove the mixed contrast objective. 
The performance of the model shows a notable degradation, especially for the knowledge relevance metric, i.e., KF1. 
This suggests that the proposed mixed contrast objective is effective in increasing the relevance of responses.

\noindent\textbf{No $\mathcal{Q}_{\text{Neg}}(*)$} -- 
We remove the hard negative sampling process and use randomly sampled instances as negatives.
The effectiveness of the hard negative sampling is evidenced by the decrease in the metric on KF1.

\noindent\textbf{No $\mathcal{Q}_{\text{Model}}$} -- 
We remove the negatives generated by the model.
The results suggest that model-generated negatives provides harder negative examples for the model, i.e., the knowledge that is more likely to be confounded by the LMs.

\noindent\textbf{No $\mathcal{L}_{\text{LM}}$} -- 
We remove the LM loss.
The effect of the model is a decline, especially on unseen topics. 
This results suggests that LM loss is instrumental in suppressing the catastrophic knowledge forgetting problem of the LMs in conversations~\citep{Chen2020RecallAL}.

\noindent\textbf{Only $\mathcal{L}_{\text{MLE}}$} --
This variant optimizes the model only by MLE loss.
We observe a substantial performance drop, especially on KF1, which demonstrates the effectiveness of \model in improving the knowledge relevancy and factuality of LMs.


\begin{figure}[t]
 \centering
 \includegraphics[width=1.0\columnwidth]{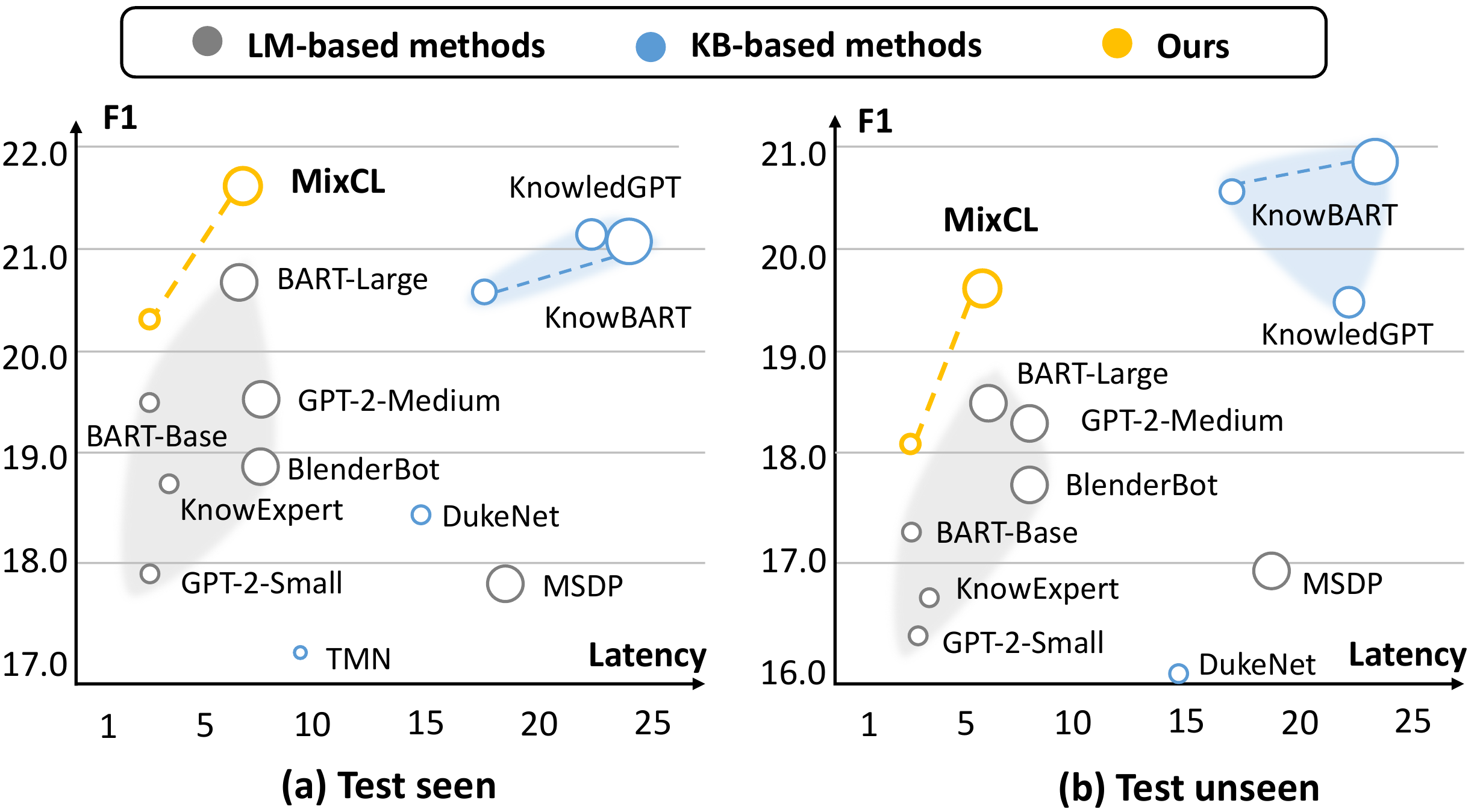}
\caption{Latency (minutes) versus response quality (F1 score) on WoW test seen and test unseen. \textcolor{gray}{Gray}, \textcolor{blue!50}{blue}, and \textcolor{Dandelion}{orange} indicate \lm, \kb, and the proposed methods, respectively. The size of the circle indicates the number of parameters of these methods.
 }
\label{fig:speed}
\end{figure}

\subsection{Efficiency analysis}
In Fig.~\ref{fig:speed}, we compare \model against baselines in terms of efficiency and effectiveness.
We adjust the inference efficiency of the models by evaluating the model with different numbers of parameters (e.g., 140M and 400M).
Compared with \kb methods, \lm methods generally have an advantage in terms of speed as they get rid of the extra IR step.
However, previous \lm methods are outperformed by \kb methods regarding response quality.
By explicitly eliminating the hallucinations of LM in conversations, \model significantly improves the response quality of \lm methods without compromising efficiency.
Notably, \model is 5$\times$ more efficient than state-of-the-art \kb methods while achieving competitive response generation performance.
Moreover, the improvements of \model along with the model size are more noticeable compared to \kb methods (see the dashed lines), indicating its superior ability to utilize the knowledge of pre-trained model.

\subsection{Case study}
We conduct several case studies and find that \model is more effective at incorporating knowledge and generating more engaging and human-like responses than baselines.
Details about our case studies are available in Appendix~\ref{sec:case} or \url{ https://github.com/sunnweiwei/MixCL}.


\section{Conclusions}
In this paper, we have proposed \model, a contrastive learning framework aimed at reducing the hallucination of language models in conversations.
\model is enhanced by negative sampling and mixed contrastive objective.
Experiments on the Wizard-of-Wikipedia dataset have shown that \model outperforms existing \lm methods and achieves comparable performance as state-of-the-art \kb methods.
Human evaluation and ablative experiments also confirm \model's effectiveness in eliminating hallucination of LMs.
Moreover, \model demonstrates advantages in terms of efficiency and scalability.
Hence, we believe that \model provides new insights on using knowledge inside large language models' parameters for KGD tasks.

The limitations of this work include the problem of knowledge forgetting.
In future work, we would like to explore practical approaches to avoiding catastrophic knowledge forgetting.
We also plan to reproduce our findings for other, less resource-rich languages.

\section*{Acknowledgements}
This work was supported by the National Key R\&D Program of
China with grant No. 2020YFB1406704, the Natural Science Foundation of China (62272274, 62202271, 61902219, 61972234, 62072279, 62102234), the Natural Science Foundation of Shandong Province (ZR2021QF129),
the Key Scientific and Technological Innovation Program of Shandong Province (2019JZZY010129), and by the Hybrid Intelligence Center, a 10-year program funded by the Dutch Ministry of Education, Culture and Science through the Netherlands Organisation for Scientific Research, \url{https://hybrid-intelligence-centre.nl}.

All content represents the opinion of the authors, which is not necessarily shared or endorsed by their respective employers and/or sponsors.

\fontsize{9.5pt}{10.5pt}
\selectfont
 
\bibliography{references}
\clearpage

\begin{appendices}
\appendix


\section{Results of pilot experiment}
\label{sec:pilot}
\begin{figure}[t]
 \centering
\includegraphics[width=1.0\columnwidth]{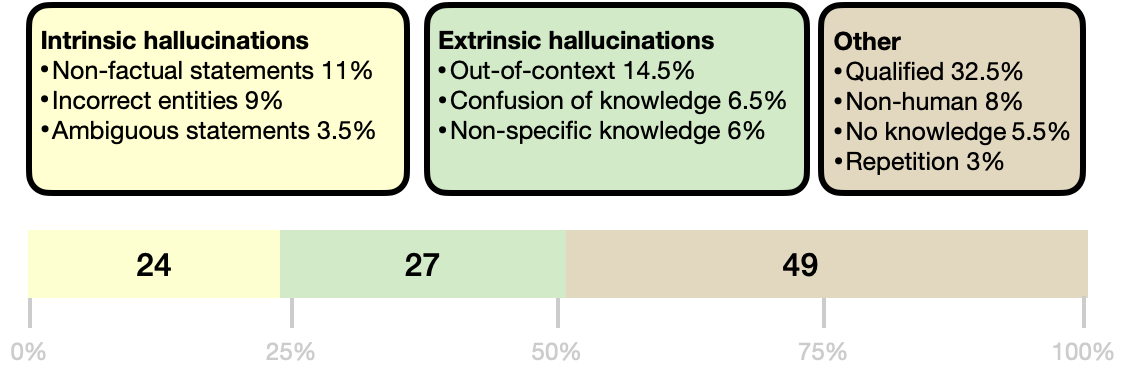} 
\caption{
Detailed results of the pilot annotation experiment.
}
\label{fig:hallucination}
\end{figure}

Our pilot human evaluation revealed that 51\% of the responses of BART are hallucinatory on the datasets of Wizard-of-Wikipedia.
We group the hallucinations into two types: \emph{intrinsic hallucinations} and \emph{extrinsic hallucinations}. The rest is categorized as \emph{other}.
Here, we provide more detailed results and examples from each type:
\begin{itemize}[leftmargin=*]
    \item \textbf{Intrinsic hallucinations} that account for 24\% of the annotated data, are non-factual statements. Specifically, intrinsic hallucinations can be further grouped into three cases.
    \begin{enumerate*}[label=(\roman*)]
    \item \emph{Non-factual statements}: we find that 11.5\% of the annotated examples made a non-factual statement, for example, ``\emph{Deer hunting is illegal in the United States.}''
    \item \emph{Incorrect entities}: we find errors in 9\% of the examples regarding entities such as time and place; for example, the model said ``\emph{Jane Eyre is a romantic novel by Charlotte Bronte.  It was published in 1891.}'' but actually, \emph{Jane Eyre} was published in 1847.
    \item \emph{Ambiguous statement}: we find that 3.5\% of the examples had ambiguous statements like ``\emph{rap lyrics are often delivered in rhyme or in punctuation.}''
    \end{enumerate*}
    
    \item \textbf{Extrinsic hallucinations} that account for 27\% of the annotated data, are irrelevant responses, which can be further grouped into three cases.
    \begin{enumerate*}[label=(\roman*)]    
    \item \emph{Out-of-context}: we find that 14.5\% of the annotated examples are out-of-context in the sense that BART uses knowledge that is inconsistent with the knowledge expected from the dialogue context, for example, a description of the history of football when the user asks for the number of teams currently in the NFL. 
    \item \emph{Confusion of similar knowledge}: we find that 6.5\% of the annotated examples suffer from confusion of knowledge where, for example, BART said ``\emph{Leonardo DiCaprio is probably the most well-known classical music composer. He was born in the city of Rome in the 15th century.}'', in which BART confuses the actor \emph{Leonardo DiCaprio} with a classical music composer.
    \item \emph{Non-specific knowledge}: we find that in 6\% of the annotated examples BART started the response by ``\emph{I don't know, but I do know that}'' and followed with a non-specific knowledge about the current dialogue topic.
    \end{enumerate*}    
    
    \item \textbf{Others} that account for 49\%. First of all, 32.5\% of the annotated examples were considered qualified responses. We find that of the remaining responses 8\% used knowledge in a non-human-like way resulting in mechanical responses, 5.5\% did not use knowledge, and 3\% repeated the knowledge used in the previous dialogue turns.
\end{itemize}
We summarize these findings in Fig.~\ref{fig:hallucination}.

\section{Implementation details}\label{sec:implementation}
\changed{We set the hyperparameters by pilot experiments following the common practice: First, we set the ratio of the three losses to \{0.4,0.3,0.3\} to make them balanced and not oversampled; Second, we use the objective shifting method (Chen et al., 2020) to vary the scale linearly to \{0.5,0.5,0\} to avoid knowledge forgetting and balance the upstream/downstream objectives.}
We set the ratio of language model loss $\alpha_3$ to $0.3$ at initialization and linearly decay it until $0$.
We set $\alpha_1$ and $\alpha_2$, i.e., the ratio of MLE loss and MCL loss, to $0.4$ and $0.3$, respectively, and linearly increase them to $0.5$ and $0.5$.
As for $\beta$, which controls the ratio of the two span extraction functions, we set it to $0.5$.
The number of negative example ($M$) is set to $8$.
We truncate the input sequence to a maximum of 128 tokens, and the output sequence to 64 tokens.
We append task-specific prompts (i.e., ``\textit{Response generation}'', ``\textit{Knowledge identification}'', and ``\textit{Wikipedia denosing}'') to three loss (i.e., $\mathcal{L}_{\text{MLE}}$, $\mathcal{L}_{\text{MCL}}$, and $\mathcal{L}_{\text{LM}}$) to distinguish different training targets separately.

All models have been trained with the AdamW optimizer with a constant learning rate of $2e{-}5$. 
We use gradient clipping with a maximum gradient norm of $0.1$. 
We optimize the model for up to 5 epochs with a batch size of 16 on 4 V100 GPUs.
\changed{During training, we first extracted spans in negatives offline, which takes about 1 hour using 20 threads. Then, we use 8 threads for getting the training samples (num\_workers=8 in data\_loader) and find that this setup does not increase the training time.}
We choose the model checkpoints by evaluating the metrics on the validation set for each 5,000 training steps.
During inference, the responses are decoded using a greedy search.
On a single GPU, the training of the model can be completed within 10h, and the average latency of the model inference for one example is 0.05s.

\begin{table*}[!t]
\small \centering
\setlength\tabcolsep{4pt}
\caption{An example of model outputs on the Wizard-of-Wikipedia test seen dataset. This example demonstrates \model's ability to improve the informativeness of a response through the use of knowledge (highlighted with \textcolor{Green}{green}).} \label{table:outputs1}

\begin{tabular}{@{}rp{15cm}@{}}

\toprule
\textbf{Topic} &{Yoga}\\
\midrule
\multirow{2}{*}{\textbf{Context}}
& \underline{User1}: Do you like yoga which is the practice not only physical but mental exercises or disciplines.\\
& \underline{User2}: I love yoga! I actually practice hot yoga pretty recently, it combines my two favorite things: Sauna and Yoga.\\
\midrule
\multicolumn{2}{@{}l}{\textbf{Model outputs}}\\
DukeNet &{i love hatha yoga too ! i like hatha yoga and raja yoga .}\\
KnowledGPT &{i love hatha yoga and raja yoga.  i like both.}\\
KnowExpert &{i love hot yoga too, it is a type of yoga that emphasizes physical exercises over emotional ones}\\
MSDP &{Yes ! It 's very intense , classes are usually an hour and a half and by the first half hour you are feeling it , the positions can be intense .}\\
\textbf{MixCL} &{I love hot yoga as well. \textcolor{Green}{It's been around since the 5th and 6th centuries BCE.}}\\
\midrule
\textbf{Human} &{Oh wow that sounds interesting. I read that Hot Yoga is actually refered to as Bikram Yoga}\\
\bottomrule
\end{tabular}
\end{table*}

\begin{table*}[!t]
\small \centering
\setlength\tabcolsep{4pt}
\caption{An example of model outputs on the Wizard-of-Wikipedia test unseen dataset. This example shows that \model uses knowledge correctly (highlighted with \textcolor{Green}{green}) but repeatedly asks ``\emph{What's your favorite store?}'' (highlighted with \textcolor{red}{red}).} \label{table:outputs2}
\begin{tabular}{@{}rp{15cm}@{}}
\toprule
\textbf{Topic} &{Online shopping}\\
\midrule
\multirow{3}{*}{\textbf{Context}}
& \underline{User2}: I like to shop online, probably a bit too much.\\
& \underline{User1}: Lol me too! \textcolor{red}{What's your favorite online store?}\\
& \underline{User2}: I love to shop on amazon. So you shop online a lot too? Where do you shop? \\
\midrule
\multicolumn{2}{@{}l}{\textbf{Model outputs}}\\
DukeNet &{i shop in amazon . i shop at the amazon . they are an american multinational technology company based in seattle , washington .}\\
KnowledGPT &{i shop online at amazon. it's a great online store. i like to shop at amazon.}\\
KnowExpert &{i shop online at amazon. i like to shop online because i can find the best deals online.}\\
MSDP &{Yes I do . I shop online a lot too . I know that Online shopping is a form of electronic commerce which allows consumers to directly buy goods or services from a seller over the Internet using a web browser .}\\
\textbf{MixCL} &{I do! I like to shop online at \textcolor{Green}{Amazon.com and Barnes \& Noble.} \textcolor{red}{What's your favorite store?}}\\
\midrule
\textbf{Human} &{Amazon as well. I also shop a lot online at Old Navy and Gap. They have great sales.}\\
\bottomrule
\end{tabular}
\end{table*}

\begin{table*}[!t]
\small \centering
\setlength\tabcolsep{4pt}
\caption{An example of model outputs on the Wizard-of-Wikipedia test unseen dataset. This example shows \kb methods failing due to a retrieval error (highlighted with \textcolor{red}{red}). \model generates a desired response that answers the user's question and incorporates knowledge (highlighted with \textcolor{Green}{green}).} \label{table:outputs3}
\begin{tabular}{@{}rp{15cm}@{}}
\toprule
\textbf{Topic} &{Archery}\\
\midrule
\multirow{3}{*}{\textbf{Context}}
& \underline{User2}: So I picked up archery recently and to my surprise, I really enjoy it. Have you ever tried?\\
& \underline{User1}: Yes, actually! I enjoy archery very much. I actually found out it comes from (arcus) Latin!\\
& \underline{User2}: What other Latin words do you know? \\
& \underline{User1}: I don't know many, but I do know that a significant portion of English comes from Latinate sources. \\
& \underline{User2}: Huh. When did you start archery? Do you take part in any competitions? \\
& \underline{User1}: I started when I was ten. I got a winning streak when I was 15, which means I won a consecutive sequence of games. \\
& \underline{User2}: I think I want to enter one but I have no idea where to start. \\
\midrule
\multicolumn{2}{@{}l}{\textbf{Model outputs}}\\
DukeNet &{the earliest evidence of use traces to the olmecs ( mexico ) , with evidence of \textcolor{red}{chocolate beverages} dating to 1900 bc}\\
KnowledGPT &{you should start by \textcolor{red}{playing chess}. it's played by millions of people worldwide.}\\
KnowExpert &{i'm not sure, but i do know that archery is a sport that involves using a bow and arrow to shoot a long-range shot at targets on a moving target.}\\
MSDP &{I think I want to enter one but I have no idea where to start .}\\
\textbf{MixCL} &{You should start with the \textcolor{Green}{World Archery Championship. It's the biggest one in the world.}}\\
\midrule
\textbf{Human} &{You could enter the Junior Grand Prix! It's a series of international junior level competitions.}\\
\bottomrule
\end{tabular}
\end{table*}

\begin{table*}[!t]
\small \centering
\setlength\tabcolsep{4pt}
\caption{An example of model outputs on the Wizard-of-Wikipedia test unseen dataset. This example shows \model producing a non-factual response (highlighted with \textcolor{red}{red}).} \label{table:outputs4}
\begin{tabular}{@{}rp{15cm}@{}}
\toprule
\textbf{Topic} &{Paramedic}\\
\midrule
\multirow{3}{*}{\textbf{Context}}
& \underline{User2}: my friend wanted to be a paramedic\\
& \underline{User1}: I wouldn't want to be EMS on 911 calls.  The scenes would haunt you for life.\\
& \underline{User2}: yea thats why my friend quit \\
& \underline{User1}: I imagine it is not as bad not that there is enchanced 911 so paramedics only have to go to the calls they are absolutely neccessary for  \\
& \underline{User2}: oh yea really? \\
& \underline{User1}: In Winnipeg they have was is known as the winnipeg fire paramedic servic, or WFPS.  6They have 2 different paramedic divisions depending on the type of 911 call. \\
& \underline{User2}: oh i didnt know that.\\
& \underline{User1}: it started in 1882 with the Winnipeg Fire Department and grew from there.\\
& \underline{User2}: dang thats so long ago\\
\midrule
\multicolumn{2}{@{}l}{\textbf{Model outputs}}\\
DukeNet &{it was built in 1738 the first fort in 1738 .}\\
KnowledGPT &{the library has both adult and children's books in over 30 languages. i think i have a lot of books in my library.}\\
KnowExpert &{it was founded in 1881 by a man named william c. durant.}\\
MSDP &{Wow that's so long ago . I know that Paramedic is a children 's book by Dr. Seuss , first published on August 12 , 1960 .}\\
\textbf{MixCL} &{Yes, the Winnipeg Fire Department is \textcolor{red}{one of the largest} fire departments in the world.}\\
\midrule
\textbf{Human} &{It is from our view point but  not in the grand scheme of things.}\\
\bottomrule
\end{tabular}
\end{table*}

\begin{table*}[!t]
\small \centering
\setlength\tabcolsep{4pt}
\caption{An example of model outputs on the Wizard-of-Wikipedia test seen dataset. This example shows \model using distinct knowledge (highlighted with \textcolor{Green}{green}) from human, but it is also appropriate.} \label{table:outputs5}
\begin{tabular}{@{}rp{15cm}@{}}
\toprule
\textbf{Topic} &{Valedictorian}\\
\midrule
\multirow{3}{*}{\textbf{Context}}
& \underline{User2}: I hope I am Valedictorian of my graduating class.\\
& \underline{User1}: Valedictorian is an academic title of success used in the United States, Canada, Central America, and the Philippines for the student who delivers the closing or farewell statement at a graduation ceremony\\
& \underline{User2}: I'm not sure what kind of grades I'll need.\\
\midrule
\multicolumn{2}{@{}l}{\textbf{Model outputs}}\\
DukeNet &{the chosen valedictorian is the highest ranking grade average}\\
KnowledGPT &{the chosen valedictorian is often the student with the highest ranking (highest grade point average, or gpa for short) among their graduating class}\\
KnowExpert &{valedictorian is also known as a bachelors of arts or bsc}\\
MSDP &{I don't know what kind of grades I 'll need . I know that: The valedictory address generally is considered a final farewell to classmates , before they disperse to pursue their individual paths after graduating .}\\
\textbf{MixCL} &{In the United States, the title is sometimes awarded to a member of a graduating university class \textcolor{Green}{on the basis of contribution to the school rather than academic success.}}\\
\midrule
\textbf{Human} &{valedictorian is often the student with the \textcolor{Green}{highest ranking} (highest Grade Point Average, or GPA for short) among their graduating class.}\\
\bottomrule
\end{tabular}
\end{table*}
\section{Case study} \label{sec:case}
We provide examples of model output in Table~\ref{table:outputs1}, \ref{table:outputs2}, \ref{table:outputs3},\ref{table:outputs4}, and \ref{table:outputs5}. 
In Table~\ref{table:outputs1}, we see that \model generates an informative response through the use of knowledge (i.e., the history of yoga).
Compared to the baselines, \model's response is more specific and relevant: DukeNet and KnowledGPT express subjective preferences rather than knowledge, KnowExpert repeats the knowledge from the previous round, and MSDP's response is hallucinating.

In Table~\ref{table:outputs2}, we see that despite appropriate integration of knowledge, \model repeatedly asked ``\emph{What's your favorite store}'' as it ignores the dialogue context, leading the annotator to consider its response is non-human.
Still, we find \model's response to be competent compared to the baselines.
Besides, we find that MSDP's response, although very informative, is wordy and preachy with its rigid insertion of large paragraphs of knowledge.

In Table~\ref{table:outputs3}, we see that \kb methods (DukeNet and KnowledGPT) suffer from the problem of retrieval error.
Their knowledge retrieval module misinterpreted the dialogue context and retrieved irrelevant knowledge (e.g., chocolate beverages and chess).
Meanwhile, we find that MSDP simply repeats the last utterance.
In contrast, \model's response is the best in terms of relevance, factuality, and humanlikeness, because it responds to the user's question, uses the correct knowledge, and is fluent in expression.

In Table~\ref{table:outputs4}, we provide an example that \model produces a non-factual response. In Table~\ref{table:outputs5}, we provide an example that \model uses appropriate knowledge, albeit distinct from the knowledge in the human response.

\end{appendices}
\end{document}